\documentclass{article}
\usepackage[T1]{fontenc}
\usepackage{iclr2027_conference,times}
\iclrfinalcopy
\usepackage{amsmath,amssymb,amsthm,booktabs,graphicx,xcolor,tabularx}
\usepackage{hyperref}
\usepackage{url}
\usepackage{multirow}
\usepackage{needspace}
\definecolor{heirblue}{RGB}{25,72,104}
\hypersetup{colorlinks=true,linkcolor=heirblue,citecolor=heirblue,urlcolor=heirblue}
\newcommand{\heir}{\textsc{HEIR}}
\newcommand{\rift}{CoRISP}

\title{HEIR: Learning Human-Entity Interactions with Functional Roles}
\author{\parbox[t]{\dimexpr\textwidth-2\tabcolsep\relax}{\raggedright\normalfont
Di Wen\textsuperscript{1,*}, Wenhao Guo\textsuperscript{1,*}, Yuedong Tan\textsuperscript{2}, Yun Huang\textsuperscript{1},\\
Minheng Wu\textsuperscript{1}, Zhihang Chen\textsuperscript{1}, Haiwen Sun\textsuperscript{1}, Fei Teng\textsuperscript{3},\\
Zhiyuan Gao\textsuperscript{4}, Yufeng Zhang\textsuperscript{1}, Yuanhao Luo\textsuperscript{1}, Jingqi Zhang\textsuperscript{1},\\
Yufan Chen\textsuperscript{1}, Junwei Zheng\textsuperscript{5}, Ruiping Liu\textsuperscript{1}, Jiale Wei\textsuperscript{1},\\
Kailun Yang\textsuperscript{3}, Kunyu Peng\textsuperscript{1,\textdagger}\\[0.6em]
{\small
\textsuperscript{1}Karlsruhe Institute of Technology (KIT)\\
\textsuperscript{2}Institute for Computer Science, Artificial Intelligence and Technology (INSAIT)\\
\textsuperscript{3}Hunan University\quad
\textsuperscript{4}University of Bremen\quad
\textsuperscript{5}ETH Zurich\\[0.4em]
\textsuperscript{*}Equal contribution.\quad
\textsuperscript{\textdagger}Corresponding author.
}}}

\begin{document}
\maketitle
\fancyhead{}
\renewcommand{\headrulewidth}{0pt}
\begin{abstract}
Understanding human--entity interactions requires recovering each person--action event's participants, roles, and shared identities. This structure can support embodied agents by clarifying who acts on which entities and how, informing anticipation and coordination in shared environments. Standard HOI metrics score individual links, leaving complete event composition undermeasured.
We introduce \heir{} (Human--Entity Interactions with Functional Roles), an image benchmark for complete grounded participant--role sets across object, interpersonal, and self-directed interactions. It contains 18,730 images, six roles, 105 actions, and 437 nouns, with shared entities, role changes, and repeated fillers; 51.6\% of images contain multiple actors and 62.1\% contain multiple actions. HEIR pairs relation AP with complete-set AP and structural evaluation.
We also introduce \rift{} (Compositional Role-aware Interaction Set Prediction), which uses shared entity identities to combine role-conditioned evidence and predict normalized participant--role sets. Cardinality and role-multiplicity potentials couple assignments through event size and role composition, with exact per-event normalization. Across 16 baselines, relation and complete-event rankings diverge even after aligning action weights. CoRISP leads the evaluated systems on repeated-role events and shared-participant images in HEIR by 2.87 and 3.82 Set mAP points, respectively. On V-COCO, CoRISP achieves 73.72/76.23 role AP and 61.06/68.59
complete-set AP on two-slot actions under Scenarios~1/2. These results show the value of learning and evaluating event composition alongside individual relations. The code and dataset are publicly available at \url{https://github.com/Kratos-Wen/HEIR}.
\end{abstract}

\section{Introduction}
A child pedals a bicycle as a parent runs alongside, steadying the saddle. A detector may recognize the child riding the bicycle yet fail to recover the event structure that tells us whether the child is riding unaided: the bicycle supports the child in the ride and is the target of the parent’s stabilizing action. Understanding the scene therefore requires more than detecting plausible person–action–object relations. A model must recover the complete set of grounded participants in each person’s action, assign each participant its functional role, and preserve entity identity across concurrent events. For embodied AI~\citep{chang2025partnr,yuan2024robopoint,li2023behavior}, this role- and identity-aware grounding can inform action planning: a robot that knows who is acting on which entity—and whether that entity functions as a target, instrument, or support—can better anticipate human activity and coordinate its actions with people in shared environments. \textit{Can a vision model do this reliably in a single image, including when several participants share a role, people act on one another, or actions are self-directed?}

Most human–object interaction evaluations emphasize localized person--action--object relations scored individually~\citep{gkioxari2018interactnet,chao2018hico}. Such scores can reward a correctly detected relation without showing whether all participants have been assembled into the right event or whether shared entities have been tracked consistently across events. We address this gap with HEIR, a benchmark that evaluates complete participant–role sets for each person–action event, and CoRISP, which predicts a normalized distribution over participant--role sets while preserving entity identities across actions.

The benchmarks that shaped image-based interaction research were designed for complementary tasks. V-COCO grounds objects in action-specific roles \citep{gupta2015vcoco}; HICO-DET localizes person--verb--object triplets over the COCO object vocabulary \citep{chao2018hico}; and SWiG grounds semantic arguments of a salient activity \citep{pratt2020swig}. A correct local prediction can receive credit while another actor's relation to the same entity is missed. Existing benchmarks~\citep{gupta2015vcoco,chao2018hico} penalize that omission when it is annotated, but credit for the first relation does not establish that the concurrent actions have been recovered together. Meanwhile, work on social interaction and open-vocabulary HOI has brought new settings and vocabularies into focus~\citep{wei2024nvi,lei2025inpcc}. Large pretrained vision--language models can recognize a wider range of interactions, but CrossHOI-Bench finds that they still struggle to assign concurrent actions to the correct person \citep{lei2026crosshoi}. Recognition of an action or noun therefore does not establish the composition of the event in which it appears.

We introduce \heir{} as a benchmark for reconstructing interaction composition in a single image. Its unit is the person--action event: a complete, variable-sized set of localized participants, each assigned a functional role, with entity identity preserved wherever it recurs. A shared ontology spans six roles, \textit{i.e.}, target, instrument, support, source, destination, and constraint, across 105 actions and 437 nouns from everyday and specialized settings. Participants include objects, other people, and localized body regions in self-directed actions; entities may recur across events in different roles, and several entities may fill the same role within one event. By bringing concurrent actors, shared entities, role changes, and repeated fillers into one annotation protocol, \heir{} makes these sources of compositional complexity directly measurable. Across its 18,730 images, 51.6\% contain at least two acting people, 62.1\% contain at least two actions, and 4,178 person--action events contain multiple fillers for a role. \heir{} measures relation and complete-event AP, exposing composition errors in multi-participant events, shared-entity scenes, and repeated-role events. Together, these design choices pose a focused question: can a model recover each event’s complete role structure while binding shared participants consistently across the image?

On HEIR, standard HOI baselines can be evaluated on role-qualified relations, but relation AP does not reveal whether all participants and roles have been assembled into the correct event. For these methods~\citep{liao2022genvlkt,yuan2023rlipv2,kim2025lain,lei2025inpcc,sun2026slhoi}, this distinction in evaluation is especially consequential for shared entities and repeated role fillers, where individually plausible links can still leave an event incomplete or bind its participants inconsistently. 
At the modeling level, contextual relation scores do not by themselves constrain event size or role multiplicity. This motivates coupling participant assignments through a normalized set distribution while preserving the identities of entities shared across actions.

We develop \rift{} to predict this set-level target directly. It uses image-level entity identities to exchange role-specific summaries across three linked neighborhoods: participants within an event, actions involving the same person--entity pair, and events sharing an entity. For each person--action event, \rift{} defines a normalized distribution over assignments of candidate entities to functional roles. Cardinality and role-multiplicity potentials couple participant assignments through event size and repeated role fillers. A dynamic program computes the exact partition function and role marginals for each event over retained proposals and fixed support. The resulting model aligns its output with \heir{}’s evaluation target while representing uncertainty over participants and set composition.

Our contributions are threefold:

\begin{enumerate}
    
\item\noindent\textbf{An event-complete benchmark.} HEIR jointly evaluates complete grounded participant--role sets for each annotated person--action event across object, interpersonal, and self-directed interactions. Its annotations preserve entity identity across concurrent events, even when an entity changes roles, and allow multiple participants to fill the same role. HEIR thus makes shared-entity reasoning and event composition explicit, measurable challenges.

\item\noindent\textbf{A composition-aware evaluation.} HEIR pairs role-qualified relation AP with complete-set AP and reports results for single- and multi-participant events, shared-entity images, and repeated role fillers. This evaluation distinguishes detecting individual relations from correctly binding participants to actors and roles and recovering each event as a complete set.

\item\noindent\textbf{A normalized model of interaction sets.} CoRISP predicts a normalized distribution over each event’s complete participant--role assignments. It uses shared entity identities to aggregate role-specific evidence within an event, across actions involving the same person--entity pair, and across events that share an entity. Cardinality and role-multiplicity potentials couple assignments through event size and repeated role fillers. A dynamic program computes exact per-event normalization and role marginals over retained proposals and fixed support.
\end{enumerate}

\section{Related Work}
\noindent\textbf{Human--Object Interaction Detection.}

Human--object interaction (HOI) detection spans appearance-, geometry-, and graph-based recognition of localized human--verb--object relations \citep{gkioxari2018interactnet,qi2018gpnn,ulutan2020vsgnet,gao2020drg} as well as query-based interaction prediction (HOTR, QPIC) and cascaded detection and interaction decoders (CDN) \citep{kim2021hotr,tamura2021qpic,zhang2021cdn}. STIP, MUREN, and PViC strengthen interaction proposals and relational or predicate-specific context \citep{zhang2022stip,kim2023muren,zhang2023pvic}. SWiG-HOI broadens object coverage~\citep{wang2021swighoi}; vision--language and open-vocabulary methods extend the interaction vocabulary~\citep{liao2022genvlkt,yuan2023rlipv2,kim2025lain,lei2025inpcc,sun2026slhoi}, while RoHOI evaluates robustness to image corruptions~\citep{wen2025rohoi}. Across these advances, standard HOI protocols chiefly score localized relations, leaving unclear whether every participant in each person--action event has been recovered, assigned the right role, and tracked consistently across concurrent interactions. HEIR makes complete grounded participant--role sets the unit of an image-level benchmark, spanning object, interpersonal, and self-directed actions while preserving shared identities and allowing repeated role fillers. CoRISP directly models this target through role-preserving cross-event context and a normalized set distribution with cardinality and role-multiplicity potentials. A dynamic program computes exact per-event normalization over retained candidates and fixed support.

\noindent\textbf{Grounded Roles and Interaction Composition.}
Prior work has advanced grounded interaction understanding along complementary directions. V-COCO~\citep{gupta2015vcoco} annotates action-specific participant roles, while SWiG~\citep{pratt2020swig} grounds the semantic arguments of a salient activity and evaluates whether its role values and boxes are jointly recovered. NVI-DET~\citep{wei2024nvi} models nonverbal behavior among individuals and groups, and GroupHOI~\citep{hong2025grouphoi} uses geometric and semantic group context to improve pairwise HOI prediction. HOI-M3~\citep{zhang2024hoim3} captures multi-person, multi-object interactions in 3D sequences for reconstruction and generation. CrossHOI-Bench~\citep{lei2026crosshoi} evaluates multi-person attribution through multiple-answer and multiple-choice questions, including image-wide settings; \citet{luo2026rethinking} jointly study pair-centric detection and anticipation in video. We use still images to isolate within-scene participant--role composition from temporal tracking and anticipation. Building on these complementary efforts, HEIR centers evaluation on each grounded person--action event in an image: it scores the complete participant--role set, allows multiple fillers per role, and preserves entity identity across concurrent events, including when an entity's role changes.

\section{The \heir{} Benchmark}
\label{sec:heir}
\noindent\textbf{Grounded events.} HEIR represents each annotated person--action event as a participant--role set in a still image. Entity $j$ has one image-level identity, noun $n_j$, and box $b_j$, reused across events. For person $h$, action $v$, and directed relations $\mathcal Q_I$, the event is $S_{h,v}=\{(j,r):(h,j,v,r)\in\mathcal Q_I\}$. Participants include objects, other people, and separately localized body regions in self-directed actions. Each participant has one role per event; multiple entities may fill the same role, and an entity may change roles across events.

\noindent\textbf{Functional roles.} The six roles describe an entity's function in an event: \texttt{target} is acted on or attended to; \texttt{instrument} is used to perform the action; \texttt{support} bears weight or supports posture; \texttt{source} is an origin; \texttt{destination} is an endpoint; and \texttt{constraint} restricts movement or state. Role annotation follows the depicted action and the participant's function. Definitions and annotation rules appear in Appendix~\ref{app:heir-details}.

\noindent\textbf{Construction and review.} We selected 18,730 of 220,362 images and frames from 29 sources (Appendix Table~\ref{tab:heir-sources}). Screening removes duplicates and prioritizes visible, localizable interactions with semantic and structural diversity. Every released image has complete human-corrected interaction annotations, covering participant identities, boxes, actions, and functional roles. Appendix~\ref{app:heir-quality} details model assistance, human review.

\begin{figure}[t]
\centering
\includegraphics[width=0.98\linewidth]{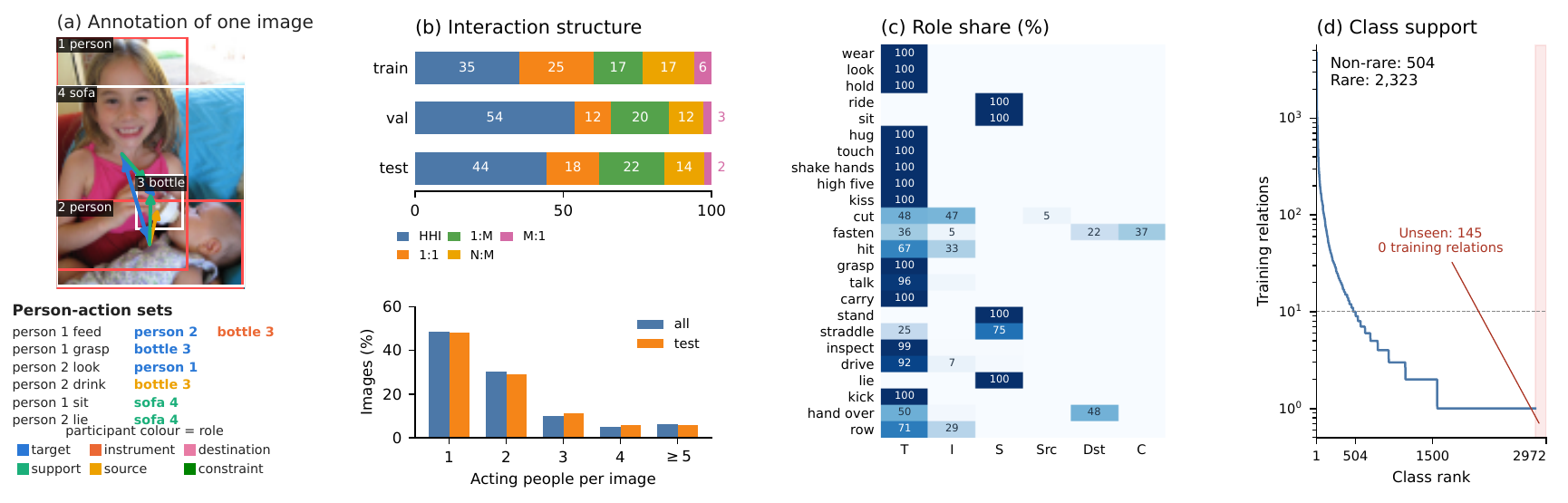}
\caption{\textbf{HEIR structure.} (a) Shared participants across events. (b) Interaction structure (top): HHI denotes images with at least one human--human relation; other images are grouped by acting-person count versus distinct participant count. Bottom: actor counts. All split images are included. (c) Role shares for the 24 most frequent actions. (d) Training support of 2,972 observed classes, including 145 unseen classes with zero training relations.}
\label{fig:heir-overview}
\end{figure}

\noindent\textbf{Interaction structure.} HEIR contains 78,345 boxes and 68,803 relations over 105 actions, 437 nouns, and six roles. Images with multiple actors and multiple actions comprise 51.6\% and 62.1\%, respectively. Cross-actor sharing occurs in 3,797 images, and 4,178 events have repeated role fillers. Figure~\ref{fig:heir-overview} illustrates these structures. The train/validation/test split contains 15,158/615/2,957 images; all primary test results use the complete test split without class exclusions.

\noindent\textbf{Evaluation.} Role mAP averages AP over test-supported $(v,n,r)$ classes. A correct relation requires the action, noun, and role, and actor and participant IoU $\geq0.5$, with one-use matching. Complete-set AP instead requires the entire event. Let $\phi_I$ be a noun-compatible, one-to-one entity correspondence at IoU $\geq0.5$, shared across an image and constructed from boxes, noun labels, and prediction confidence. A predicted event is correct only if its actor and all participants have matches, its actor and action identify an unmatched annotated event, and
\begin{equation}
 \{(\phi_I(\hat j),\hat r):(\hat j,\hat r)\in\hat S_{\hat h,\hat v}\}
 = S_{\phi_I(\hat h),\hat v}.
 \label{eq:heir-set-match-main}
\end{equation}
Missing, additional, wrongly labeled, or unmatched participants make the set incorrect. Set mAP averages all-point interpolated AP over test-supported actions. Correspondence, tied scores, duplicates, and structural strata are specified in Appendix~\ref{app:experimental-details}.

\section{CoRISP: Grounded Interaction Composition}
\label{sec:method}
\begin{figure}[t]
\centering
\includegraphics[width=\linewidth]{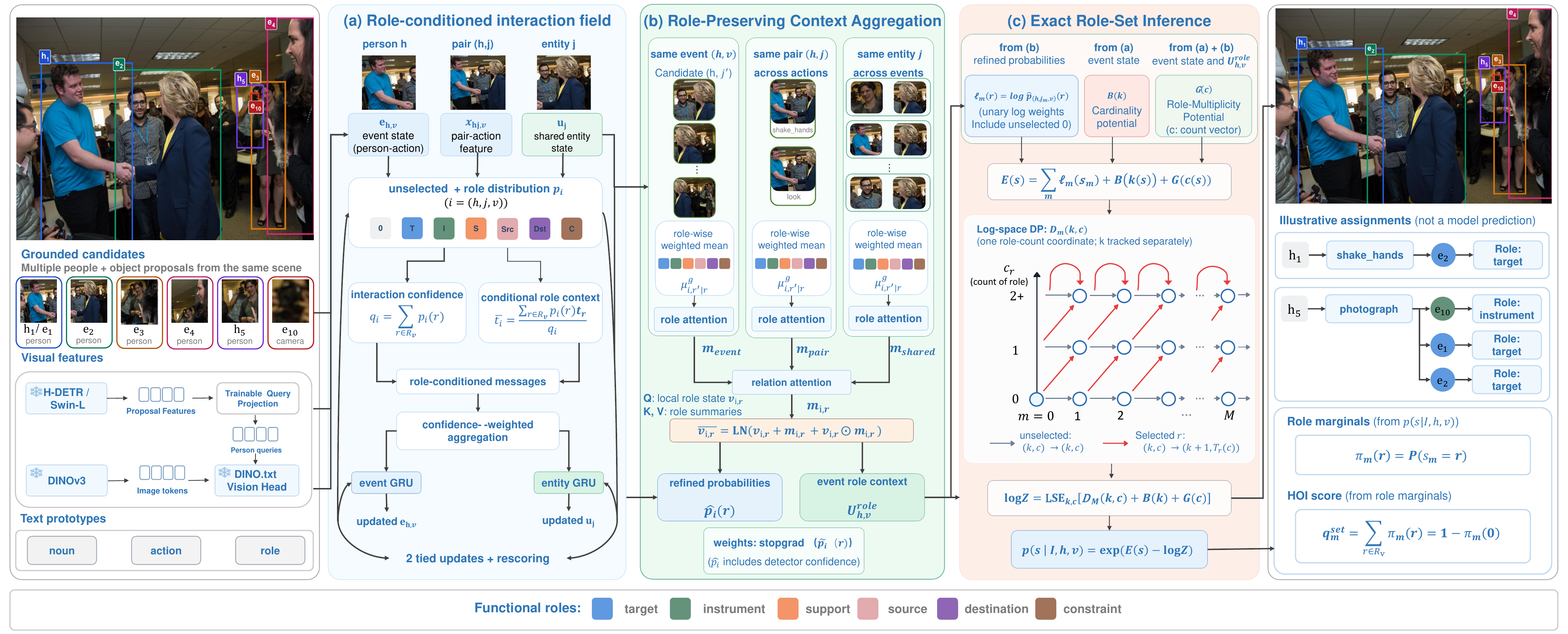}
\vskip-1em
\caption{\textbf{CoRISP overview.} (a) Role-conditioned recurrent updates. (b) Role-preserving aggregation over event, pair, and shared-entity neighborhoods. (c) Exact set normalization with cardinality and role-multiplicity potentials. Marginals score individual relations; joint probabilities score complete assignments. The displayed assignments illustrate the output structure.}
\vskip-3ex
\label{fig:rift-overview}
\end{figure}

Probabilistic set prediction models membership and cardinality~\citep{rezatofighi2017deepsetnet}. CoRISP uses participant--role assignments both to organize visual evidence and to define a normalized event distribution (Figure~\ref{fig:rift-overview}). In event $(h,v)$, each candidate takes state $0$ (unselected) or one role in $\mathcal R_v$; several candidates may share a role. Visual--text features initialize event states $\mathbf e_{h,v}$, shared entity states $\mathbf u_j$, and pair--action features $\mathbf x_{hj,v}$. Entities retain their identities across events, which exchange evidence but have separate normalizers.

\subsection{Role-Conditioned Interaction Field}
\label{sec:rift-field}
For a visible candidate $i=(h,j,v)$, a softmax $p_i$ over state $0$ and applicable roles gives interaction confidence and conditional role context:
\begin{equation}
 q_i=\sum_{r\in\mathcal R_v}p_i(r),\qquad
 \bar{\mathbf t}_i=\frac{\sum_{r\in\mathcal R_v}p_i(r)\mathbf t_r}{\max(q_i,\epsilon)}.
 \label{eq:rift-role-feedback}
\end{equation}
Here $\mathbf t_r$ is a projected role prototype and $\epsilon$ a numerical floor. Interaction confidence $q_i$ weights each message, while $\bar{\mathbf t}_i$ specifies its role content. Messages combine this context with pair features and visual evidence. Confidence-weighted means feed event and entity GRUs~\citep{cho2014gru}. Two tied updates followed by rescoring let role estimates change the features used for subsequent interaction decisions. Actions remain multilabel.

\subsection{Role-Preserving Context Aggregation}
\label{sec:rift-context}
Each pair--action index receives three kinds of context: co-participants in event $(h,v)$, other actions involving pair $(h,j)$, and other events involving entity $j$. These neighborhoods connect event composition, action-dependent roles, and shared participants. Detached positive-role probabilities weight candidates within each role; the query's same-role contribution is removed. Pooling within roles keeps evidence for different participant functions separate. Attention reads these summaries and fuses the three neighborhood messages into $\mathbf m_{i,r}$.

The local role feature $\mathbf v_{i,r}$ combines pair evidence with noun and action--role anchors. Binding it to the fused message gives $\bar{\mathbf v}_{i,r}=\operatorname{LN}(\mathbf v_{i,r}+\mathbf m_{i,r}+\mathbf v_{i,r}\odot\mathbf m_{i,r})$, where $\operatorname{LN}$ is layer normalization~\citep{ba2016layernorm} and $\odot$ is elementwise multiplication. A bounded residual $\delta_{i,r}$ refines detector-weighted probabilities $\widetilde p_i$:
\begin{equation}
 \widehat p_i(r)=
 \frac{\widetilde p_i(r)e^{\delta_{i,r}}}
 {\sum_{r'\in\{0\}\cup\mathcal R_v}\widetilde p_i(r')e^{\delta_{i,r'}}},
 \qquad \delta_{i,0}=0.
 \label{eq:rift-refinement}
\end{equation}
Role-wise pooling of the same bound features, combined with event and action--role features, gives $\mathbf U^{\rm role}_{h,v}$ for set composition. Appendix~\ref{app:rift-details} gives the pooling and potential parameterizations.

\subsection{Exact Set Normalization}
\label{sec:rift-set}
Fix an event with candidates $j_1,\ldots,j_M$. An assignment $\mathbf s=(s_1,\ldots,s_M)$ encodes $S=\{(j_m,s_m):s_m\ne0\}$, with unary log weights $\ell_m(r)=\log\widehat p_{(h,j_m,v)}(r)$, including $r=0$. Its total count is $k(\mathbf s)=\sum_m\mathbf1\{s_m\ne0\}$ and its saturated role counts are $c_r(\mathbf s)=\min(2,\sum_m\mathbf1\{s_m=r\})$. For an admissible assignment, CoRISP defines
\begin{align}
 E(\mathbf s)&=\sum_m\ell_m(s_m)+B(k(\mathbf s))+G(\mathbf c(\mathbf s)),\label{eq:rift-energy}\\
 p(\mathbf s\mid I,h,v)&=\exp E(\mathbf s)/Z,\qquad
 Z=\sum_{\mathbf s'\in\Omega_{h,v}}\exp E(\mathbf s').\label{eq:rift-set}
\end{align}
The Cardinality Potential $B$ scores total count from $\mathbf e_{h,v}$. The Role-Multiplicity Potential $G$ combines each contextual role vector with its count embedding and scores the pooled representation against the event state. Whereas $B$ distinguishes event sizes, $G$ also distinguishes role compositions at the same size, such as two targets versus a target and an instrument. Both vanish at the empty set. Saturating each role count at $2+$ retains the distinction between absence, presence, and repetition while permitting compact count-based inference. At set inference, the fixed inventory $C(v,n_j,r)$ defines admissible positive states in $\Omega_{h,v}$.

A log-space dynamic program accumulates log unary mass $D_m(k,\mathbf c)$ over the first $m$ candidates. Each transition leaves a candidate unselected or assigns one allowed role. The partition is
\begin{equation}
 \log Z=\operatorname{LSE}_{k,\mathbf c}
 [D_M(k,\mathbf c)+B(k)+G(\mathbf c)].
 \label{eq:rift-partition}
\end{equation}
Here $\operatorname{LSE}$ denotes log-sum-exp. Count-based inference follows established dynamic programming principles~\citep{tarlow2012cardinality}; exactness is within an event's retained candidates and support. Holding $B$ and $G$ fixed, role marginals are $\pi_m(r)=\partial\log Z/\partial\ell_m(r)$ and interaction confidence is $q_m^{\rm set}=1-\pi_m(0)$. Contextual unaries carry identity-specific evidence; $B$ and $G$ couple assignments through their composition.

Set prediction replaces log-sum-exp with maximization to obtain the highest-scoring assignment per count state, ranked by its normalized probability. Relation confidence and complete-event confidence follow from the same learned distribution.

\noindent\textbf{Set supervision.} Let $\mathcal A(y)\subseteq\Omega_{h,v}$ contain the distinct complete assignments compatible with annotation $y$, using each proposal at most once. The Localization-Marginalized Set Likelihood sums their probability:
\begin{equation}
 \mathcal L_{\rm set}(y)=\log Z-
 \operatorname{LSE}_{\mathbf s\in\mathcal A(y)}E(\mathbf s).
 \label{eq:rift-nll}
\end{equation}
Summing compatible assignments accommodates several proposals for one annotated participant without selecting an arbitrary localization target. Equivalent matching paths count once per assignment. Observed negatives target the empty set; unknown or unrepresentable positive
events are omitted from this loss. Matching and focal modulation are specified in Appendix~\ref{app:rift-training}.

\section{Experiments}
\label{sec:experiments}
\subsection{Experimental Setup}
\label{sec:experimental-protocols}
\noindent\textbf{Benchmarks.} HEIR uses 2,957 test images, 1,104 role classes, and 9,955 sets over 101 actions. V-COCO~\citep{gupta2015vcoco} uses eligible trainval images and all 4,946 test images.

\noindent\textbf{Implementation details.}\label{par:rift-inventory} CoRISP uses frozen H-DETR/Swin-L proposals~\citep{jia2023hdetr}. A DINOv3 ViT-L/16 backbone followed by a DINO.txt vision head supplies text-aligned visual features~\citep{simeoni2025dinov3,jose2025dinotxt}; both components are frozen. CoRISP trains its feature projections, interaction field, context aggregation, and set potentials, totaling 9.9M trainable parameters. Training runs for 30 epochs, with HEIR checkpoints selected by validation Role mAP. HEIR uses a benchmark-wide semantic compatibility inventory of 10,072 action--noun--role combinations. The inventory encodes whether a combination is semantically admissible and is shared by dagger-marked baselines before
their native top-100 selection and applicable NMS; other baselines retain their native support. UniHOI~\citep{yang2026unihoi} and HOI-IDiff~\citep{hui2025idiff} are reported under their published V-COCO conventions in a separate table block.

\noindent\textbf{Evaluation setting.} HEIR reports Role and Set mAP (Section~\ref{sec:heir}) and HOI mAP, which drops role labels and takes their maximum score. Pair-output baselines are converted into set hypotheses from ranked relation scores, while CoRISP predicts complete assignments from its learned set distribution. All methods use the same set matching protocol and the same 100-set image budget. The HEIR component-ablation protocol uses CoRISP's set prediction rule. Single/Multi/Repeat/Shared denote one-member events, multi-member events, repeated roles, and shared-entity images. V-COCO reports role AP ($\mathrm{AP}_{\mathrm{role}}$), excluding \texttt{point}, and our Set mAP under S1/S2; Dual averages three two-slot actions. Appendix~\ref{app:experimental-details} specifies matching, annotation scope, encoders, and conditional strata.

\subsection{Comparison with Existing Methods}
\label{sec:heir-results}\label{sec:comparison}
\noindent\textbf{HEIR.} Table~\ref{tab:mainresults} places relation detection alongside event recovery. CoRISP leads on Repeat and Shared Set mAP (8.99 and 23.80), exceeding the strongest baselines by 2.87 and 3.82 points with 9.9M trainable parameters. These strata test assigning several entities to one role and recovering events in shared-participant images. RLIPv2 Swin-L leads on overall Set mAP (22.21), while SOV-STG Swin-L leads on Multi (9.87).

\begin{table}[!t]
\centering
\caption{\textbf{HEIR (left) and V-COCO (right)} (\%). V-COCO pairs are S1/S2; re-evaluated role AP excludes \texttt{point}. Params: HEIR trainable millions. $^\dagger$: shared HEIR inventory; W/T/A/R: released weights, trained official code, adapted official code, or reimplementation. Bold/underline: best/second, excluding literature-only scores.}
\label{tab:mainresults}\label{tab:vcoco}
\begingroup
\small
\setlength{\tabcolsep}{2pt}
\renewcommand{\arraystretch}{1.04}
\newcommand{\vcpair}[2]{#1/#2}
\resizebox{\linewidth}{!}{%
\begin{tabular}{@{}l *{8}{r}@{\hspace{8pt}}rrr@{}}
\toprule
 & \multicolumn{8}{c}{\textbf{HEIR}} & \multicolumn{3}{c}{\textbf{V-COCO (S1/S2)}} \\
\cmidrule(lr){2-9}\cmidrule(l){10-12}
\textbf{Method} & \shortstack{Params\\(M)$\downarrow$} & \shortstack{HOI\\$\uparrow$} & \shortstack{Role\\$\uparrow$} & \multicolumn{5}{c}{\textbf{Set mAP}$\uparrow$} & $\mathbf{AP}_{\mathrm{role}}\uparrow$ & \multicolumn{2}{c}{\textbf{Set mAP}$\uparrow$} \\
\cmidrule(lr){5-9}\cmidrule(l){11-12}
 & & & & Full & Single & Multi & Repeat & Shared & & All & Dual \\
\midrule
\multicolumn{12}{@{}l}{\emph{Visual models}} \\
QPIC R50 \citep{tamura2021qpic}$^\dagger$ & 41.6 & 15.11 & 14.82 & 14.14 & 16.03 & 4.92 & 3.68 & 14.71 & \vcpair{58.79}{60.97}$^{\mathrm{W}}$ & \vcpair{53.87}{57.77} & \vcpair{41.64}{45.84} \\
QPIC R101 \citep{tamura2021qpic}$^\dagger$ & 60.5 & 15.57 & 15.26 & 14.84 & 16.93 & 3.86 & 3.01 & 15.10 & \vcpair{58.16}{60.65}$^{\mathrm{W}}$ & \vcpair{53.03}{57.11} & \vcpair{39.38}{44.28} \\
MUREN \citep{kim2023muren}$^\dagger$ & 75.1 & 16.38 & 16.17 & 13.98 & 16.04 & 6.57 & 4.66 & 15.75 & \vcpair{68.72}{70.97}$^{\mathrm{W}}$ & \vcpair{62.97}{67.33} & \vcpair{54.94}{60.95} \\
SOV-STG-S \citep{chen2025sovstg}$^\dagger$ & 54.1 & 15.55 & 15.21 & 13.64 & 15.67 & 2.62 & 1.70 & 13.00 & \vcpair{62.99}{64.56}$^{\mathrm{T}}$ & \vcpair{56.62}{59.53} & \vcpair{47.79}{50.02} \\
SOV-STG Swin-L \citep{chen2025sovstg}$^\dagger$ & 240.4 & 23.69 & 23.35 & \underline{18.93} & \underline{21.28} & \textbf{9.87} & 5.85 & 19.57 & \vcpair{69.87}{71.61}$^{\mathrm{T}}$ & \vcpair{63.86}{67.15} & \vcpair{55.20}{59.36} \\
SOV-STG R101 \citep{chen2025sovstg}$^\dagger$ & 87.4 & 17.11 & 16.86 & 14.58 & 16.89 & 5.33 & 1.57 & 16.43 & \vcpair{66.63}{68.21}$^{\mathrm{W}}$ & \vcpair{60.52}{63.63} & \vcpair{49.59}{53.91} \\
PViC R50 \citep{zhang2023pvic} & 12.2 & 16.24 & 16.02 & 13.83 & 16.39 & 3.93 & 4.78 & 15.17 & \vcpair{58.70}{65.88}$^{\mathrm{T}}$ & \vcpair{55.38}{64.05} & \vcpair{27.66}{41.76} \\
PViC Swin-L \citep{zhang2023pvic} & \underline{12.1} & 19.22 & 18.99 & 14.57 & 17.21 & 5.11 & \underline{6.12} & 16.62 & \vcpair{60.85}{68.72}$^{\mathrm{T}}$ & \vcpair{57.01}{66.89} & \vcpair{31.26}{49.91} \\
\midrule
\multicolumn{12}{@{}l}{\emph{Vision--language models}} \\
GEN-VLKT-S \citep{liao2022genvlkt}$^\dagger$ & 46.9 & 17.23 & 17.13 & 12.95 & 15.50 & 4.33 & 4.54 & 15.92 & \vcpair{65.05}{67.19}$^{\mathrm{W}}$ & \vcpair{59.47}{63.58} & \vcpair{50.40}{55.11} \\
GEN-VLKT-L \citep{liao2022genvlkt}$^\dagger$ & 75.4 & 19.23 & 18.92 & 15.85 & 19.34 & 4.79 & 3.25 & 17.59 & \vcpair{66.34}{68.81}$^{\mathrm{W}}$ & \vcpair{60.34}{65.11} & \vcpair{49.50}{55.43} \\
SOV-STG-VLA-S \citep{chen2025sovstg}$^\dagger$ & 88.2 & 21.60 & 21.01 & 15.45 & 19.11 & 4.01 & 1.45 & 15.33 & \vcpair{66.49}{68.46}$^{\mathrm{W}}$ & \vcpair{60.87}{64.65} & \vcpair{50.69}{55.52} \\
RLIPv2 Swin-T \citep{yuan2023rlipv2}$^\dagger$ & 213.4 & 22.19 & 21.87 & 17.52 & 20.27 & 5.40 & 3.78 & 18.09 & \vcpair{68.99}{71.12}$^{\mathrm{W}}$ & \vcpair{63.22}{67.67} & \vcpair{55.67}{62.35} \\
RLIPv2 Swin-L \citep{yuan2023rlipv2}$^\dagger$ & 382.8 & \textbf{26.38} & \textbf{25.91} & \textbf{22.21} & \textbf{25.14} & 7.26 & 4.66 & \underline{19.98} & \vcpair{\underline{71.92}}{\underline{74.20}}$^{\mathrm{W}}$ & \vcpair{66.49}{71.17} & \vcpair{58.02}{64.82} \\
GroupHOI-S \citep{hong2025grouphoi} & 82.0 & 13.68 & 13.32 & 14.52 & 16.74 & 4.22 & 3.43 & 15.24 & \vcpair{66.54}{69.54}$^{\mathrm{A}}$ & \vcpair{60.72}{66.18} & \vcpair{51.36}{58.57} \\
InCoM-Net \citep{seo2026incom} & 42.2 & 15.07 & 14.79 & 14.94 & 17.63 & 4.58 & 5.18 & 17.75 & \vcpair{70.11}{73.80}$^{\mathrm{R}}$ & \vcpair{\textbf{67.66}}{\underline{71.69}} & \vcpair{\underline{59.43}}{\underline{64.88}} \\
SL-HOI \citep{sun2026slhoi} & 31.1 & \underline{26.01} & \underline{24.92} & 10.93 & 13.48 & 1.78 & 0.27 & 10.87 & \vcpair{48.16}{50.24}$^{\mathrm{A}}$ & \vcpair{43.31}{46.77} & \vcpair{24.25}{27.50} \\
\midrule
\multicolumn{12}{@{}l}{\emph{Generative models}} \\
\label{tab:vcoco-published-generative}
HOI-IDiff \citep{hui2025idiff} & \multicolumn{8}{c}{--} & \vcpair{73.4}{76.1} & -- & -- \\
UniHOI \citep{yang2026unihoi} & \multicolumn{8}{c}{--} & \vcpair{72.91}{77.45} & -- & -- \\
\midrule
\textbf{CoRISP} & \textbf{9.9} & 21.21 & 20.41 & 18.04 & 20.57 & \underline{8.94} & \textbf{8.99} & \textbf{23.80} & \vcpair{\textbf{73.72}}{\textbf{76.23}} & \vcpair{\underline{67.11}}{\textbf{72.20}} & \vcpair{\textbf{61.06}}{\textbf{68.59}} \\
\bottomrule
\end{tabular}%
}
\endgroup
\vskip-3ex
\end{table}

\noindent\textbf{V-COCO.} Under the common evaluation, CoRISP leads in $\mathrm{AP}_{\mathrm{role}}$ (73.72/76.23) and Dual Set mAP (61.06/68.59), improving Dual by 1.63/3.71 points over the strongest baseline. It also leads on overall S2 Set mAP (72.20); InCoM-Net leads on S1 (67.66). Dual tests joint recovery of both native slots, whereas HEIR additionally tests variable cardinality and repeated roles.

\subsection{What HEIR Reveals}
\label{sec:benchmark-insights}
\begin{figure}[t]
\centering
\includegraphics[width=0.96\linewidth]{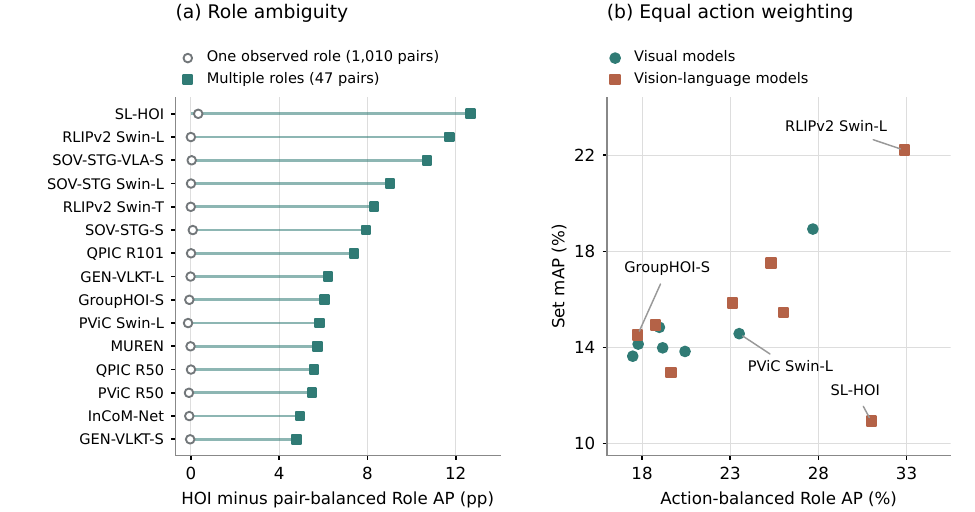}
\vskip-2ex
\caption{\textbf{Role ambiguity and event recovery across all 16 baselines.} (a) HOI--Role AP gaps with equal action--noun pair weights. (b) Role and Set AP with equal weights over the same 101 actions; sets use common-prefix decoding.}
\vskip-3ex
\label{fig:heir-diagnostics}
\end{figure}

\noindent\textbf{Role ambiguity depends on the action--noun context.} On the 47 test-supported action--noun pairs observed with multiple roles, retaining role labels reduces pair-balanced AP by 4.78--12.66 points across 16 baselines. The absolute difference is at most 0.34 on the 1,010 single-role pairs (Figure~\ref{fig:heir-diagnostics}a). Aggregate HOI--Role differences therefore mask the difficulty of assigning a participant's function when its action and noun admit several roles.

\noindent\textbf{Better relations need not yield better events.} Role mAP weights semantic classes, whereas Set mAP weights actions. After averaging Role AP within actions and then over the same 101 actions, 41 of 120 pairs among the 16 baselines still reverse order between the two metrics (Figure~\ref{fig:heir-diagnostics}b). Ten of 55 pairs reverse within the shared-inventory baselines alone. SL-HOI exceeds PViC Swin-L by 7.49 points in action-balanced Role AP but trails it by 3.64 in Set mAP. Thus, event recovery measures more than a different weighting of relation accuracy.

\noindent\textbf{Completeness requires excluding plausible extras.} For five baselines, only 53.58--61.81\% of edge-covered multi-participant events are recovered exactly (Appendix~\ref{app:benchmark-diagnostics}). Each failure has an extra member in every prefix containing all true members. Within these covered events, the bottleneck is separating required members from interleaved false edges. On V-COCO, the same models recover 98.52--100\% of fully covered two-slot events under S1. Native slots fix the role positions; HEIR additionally requires selecting a variable number of participants. Each rate conditions on that model's covered events.

\subsection{Evidence Availability and Selection}
\label{sec:oracle-analysis}
Following the distinction between localization and interaction errors~\citep{zhu2025diaghoi}, we examine whether complete visual evidence is available and correctly ranked. The oracle supplies the true action and participant count $k^\star$ at a fixed budget of 100 relations/image. Coverage $C$ is the fraction of events with noun-compatible boxes for the actor and every participant; recovery $R$ is the fraction whose top-$k^\star$ edges form the correct set (Table~\ref{tab:diagnostic-summary}; all models in Table~\ref{tab:cardinality-oracle}).

CoRISP recovers 27.48\% of Multi events and 33.41\% of Repeat events, exceeding the strongest baselines by 3.26 and 6.52 points. Its Multi coverage is 48.92\%, compared with SOV-STG Swin-L's 51.21\%, yet its recovery is higher: 27.48\% versus 24.22\%. Thus, higher candidate coverage alone does not determine which system ranks complete sets correctly. CoRISP also provides the highest Repeat coverage (62.01\%) and recovery. Coverage reflects the boxes and nouns present in the retained relations; recovery additionally requires correct member and role ranking. Together, these results identify complementary priorities: improve complete-participant coverage through localization, noun recognition, and candidate retention, while preserving CoRISP's stronger ranked-set recovery.

A complementary comparison holds the evaluated person--participant pairs fixed and requires both models to localize them with the correct noun. On 10,383 such pairs, including negative pairs, CoRISP attains 61.50 conditional Role mAP versus SL-HOI's 50.26 over the same 584 supported classes (Appendix~\ref{app:conditional-grounding}). Actions and roles remain predicted. This comparison separates classification on jointly recoverable pairs from the coverage losses measured above.

\begin{table}[!t]
\begin{minipage}[t]{0.52\linewidth}
\vspace{0pt}
\caption{\textbf{HEIR component ablations} (\%). The
model and evaluation settings follow Table~\ref{tab:mainresults}}
\label{tab:heir-ablation}\label{tab:ablation}
\begingroup
\small
\setlength{\tabcolsep}{3pt}
\resizebox{0.95\linewidth}{!}{%
\begin{tabular}{@{}l *{6}{r}@{}}
\toprule
\textbf{Variant} & \shortstack{Role\\Full$\uparrow$} & \multicolumn{5}{c}{\textbf{Set mAP}$\uparrow$} \\
\cmidrule(l){3-7}
 & & Full & Single & Multi & Repeat & Shared \\
\midrule
w/o $G$ & 20.67 & \underline{17.09} & 20.54 & 6.89 & \underline{5.39} & 19.96 \\
\shortstack[l]{w/o context\\messages} & \underline{20.93} & 16.42 & 19.40 & 6.70 & 4.38 & 19.75 \\
\shortstack[l]{w/o role\\feedback} & \textbf{20.95} & 17.05 & \underline{20.18} & \underline{6.97} & 4.17 & \underline{19.97} \\
\midrule
\shortstack[l]{CoRISP} & 20.41 & \textbf{18.04} & \textbf{20.57} & \textbf{8.94} & \textbf{8.99} & \textbf{23.80} \\
\bottomrule
\end{tabular}%
}
\endgroup
\end{minipage}\hfill
\begin{minipage}[t]{0.46\linewidth}
\vspace{0pt}
\caption{\textbf{Action--cardinality oracle} (\%). The true action and count are supplied. $C$: coverage; $R$: recovery. All use 100 relations/image; emphasis ranks shown models.}
\label{tab:diagnostic-summary}\label{tab:oracle-summary}
\begingroup
\small
\setlength{\tabcolsep}{3pt}
\resizebox{\linewidth}{!}{%
\begin{tabular}{@{}lrrrr@{}}
\toprule
\textbf{Method} & \multicolumn{2}{c}{Multi} & \multicolumn{2}{c}{Repeat} \\
\cmidrule(lr){2-3}\cmidrule(l){4-5}
 & $C\uparrow$ & $R\uparrow$ & $C\uparrow$ & $R\uparrow$ \\
\midrule
\shortstack[l]{SOV-STG Swin-L\\\citep{chen2025sovstg}} & \textbf{51.21} & \underline{24.22} & 56.06 & 24.94 \\
\addlinespace[2pt]
\shortstack[l]{PViC Swin-L\\\citep{zhang2023pvic}} & 47.47 & 22.55 & \underline{60.41} & \underline{26.89} \\
\addlinespace[2pt]
\shortstack[l]{RLIPv2 Swin-L\\\citep{yuan2023rlipv2}} & 48.30 & 23.94 & 54.58 & 25.51 \\
\midrule
CoRISP & \underline{48.92} & \textbf{27.48} & \textbf{62.01} & \textbf{33.41} \\
\bottomrule
\end{tabular}%
}
\endgroup
\end{minipage}
\vspace{-1em}
\end{table}

\subsection{CoRISP Component Analysis}
\label{sec:rift-analysis}

Removing $G$ retains cardinality scoring but removes the role-composition
potential. Removing context messages disables event, pair, and shared-entity
aggregation while retaining the recurrent field. Removing role feedback
excludes the conditional role vector from recurrent messages while preserving
confidence weighting. These removals reduce Multi Set mAP from 8.94 to 6.89,
6.70, and 6.97, respectively. Removing context messages causes the largest
drop in overall Set mAP, from 18.04 to 16.42. Interestingly, each deletion
slightly increases aggregate Role mAP while reducing Set mAP, reinforcing the
distinction between scoring individual relations and recovering complete event
composition. With the primary checkpoint fixed, retaining up to eight
count-state winners improves Set mAP from 17.56 for the single best nonempty
assignment to 18.04 (Table~\ref{tab:rift-decoding}).

\begin{figure}[t]
\noindent\input{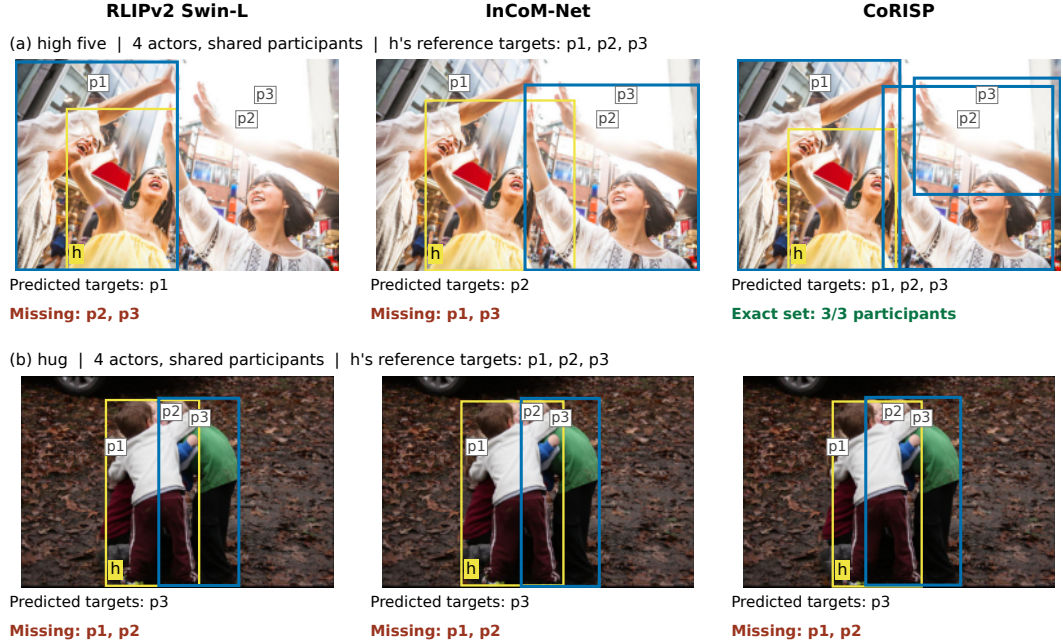}
\vskip-2em
\caption{\textbf{Composition in four-person scenes.} Highest-scoring set for the indicated actor/action, on identical uncropped images. Gray IDs mark reference participants; solid yellow/blue boxes are predicted actors/targets. In (a), CoRISP recovers three targets sharing one role; in (b), all models omit two participants in an occluded group hug.}
\label{fig:qualitative}
\vskip-3ex
\end{figure}

\noindent\textbf{Qualitative analysis.} Figure~\ref{fig:qualitative} contrasts finding a plausible partner with recovering all participants. In the four-person high-five, CoRISP ranks the complete three-target set first, while both baselines rank a singleton first. The distinction is between confidence in an individual relation and confidence that an event has no missing participants. In the group hug, all three models recover a valid target but omit two heavily occluded participants.

\section{Discussion and Limitations}
HEIR makes complete event composition measurable alongside relation detection. The ranking reversals and covered-but-incomplete events show why both views matter. CoRISP combines role-conditioned evidence with a normalized participant-set model, with its strongest system-level gains on repeated roles, shared-participant images, and V-COCO's two-slot actions. Candidate coverage remains a practical limit: an event requires suitable boxes and noun labels for every participant. The oracle results motivate improving this front end while preserving the model's stronger ranked-set recovery.
\label{maintextend}

\clearpage
\section*{Acknowledgment}
The project is funded by the Deutsche Forschungsgemeinschaft (DFG, German Research Foundation) -- SFB1574 -- 471687386. This work was supported in part by the SmartAge project sponsored by the Carl Zeiss Stiftung (P2019-01-003; 2021--2026). The authors gratefully acknowledge the computing time provided on the high-performance computer HoreKa by the National High-Performance Computing Center at KIT (NHR@KIT). This center is jointly supported by the Federal Ministry of Education and Research and the Ministry of Science, Research and the Arts of Baden-W\"urttemberg, as part of the National High-Performance Computing (NHR) joint funding program (\url{https://www.nhr-verein.de/en/our-partners}). HoreKa is partly funded by the German Research Foundation (DFG).

\subsection*{AI Use Statement}
Generative AI was used to aid and polish manuscript writing, as well as to assist with retrieval and discovery during literature review. Additionally, multimodal models were used to propose initial dataset annotations prior to human correction and review, as described in Appendix~\ref{app:heir-quality}. The authors take full responsibility for the final text, claims, code, and data, including all AI-assisted contributions.

\subsection*{Ethics Statement}
HEIR reannotates images from existing public sources, including interpersonal and clinical scenes. Source provenance is listed in Appendix Table~\ref{tab:heir-sources}. Redistribution is subject to each source's license and privacy conditions; where image redistribution is not permitted, the release will provide identifiers and acquisition references. Annotation and review coverage are described in Appendix~\ref{app:heir-quality}.

\subsection*{Reproducibility Statement}
Appendix~\ref{app:rift-details} specifies CoRISP's probabilities, potentials, dynamic program, and supervision. Appendix~\ref{app:experimental-details} defines matching, decoding, output budgets, model selection, and diagnostic protocols. The code and dataset are publicly available at \url{https://github.com/Kratos-Wen/HEIR}.

\bibliography{references}
\bibliographystyle{iclr2027_conference}
\clearpage
\appendix
\section{Annotation Provenance and Quality}
\label{app:heir-details}\label{app:data}
\noindent\textbf{Sources.} Table~\ref{tab:heir-sources} lists the 29 sources. Sources with native interactions contribute their boxes and labels as annotation evidence; other sources enter through model proposals. Each image, including a sampled video frame, is annotated independently with image-local entity identities.

\noindent\textbf{Split isolation.} Frames from the same source video or capture sequence, together with near-duplicate images, are assigned to a single split.

\begin{table}[t]
\centering
\caption{\textbf{Source datasets and collections.} Released images per source and the setting each source contributes. HICO-DET, V-COCO, NVI, and SWiG enter with native interaction labels; all other sources are annotated from model proposals.}
\label{tab:heir-sources}
\small
\setlength{\tabcolsep}{4pt}
\resizebox{\ifdim\width>\linewidth\linewidth\else\width\fi}{!}{\begin{tabular}{@{}llrl@{}}
\toprule
Setting & Source & Released & What it contributes \\
\midrule
\multirow{7}{*}{Everyday} & HICO-DET~\citep{chao2018hico} & 6,347 & Web photographs, 117 actions, COCO nouns \\
 & V-COCO~\citep{gupta2015vcoco} & 1,848 & COCO images with action-specific roles \\
 & SWiG~\citep{pratt2020swig} & 1,828 & Situations with grounded semantic roles \\
 & Visual Genome~\citep{krishna2017visualgenome} & 216 & Dense scene-graph relations \\
 & Open Images~\citep{kuznetsova2020openimages} & 191 & Visual relationship annotations \\
 & Wikimedia Commons~\citep{wikimediacommons} & 157 & Openly licensed photographs \\
 & Openverse~\citep{openverse} & 103 & Openly licensed photographs \\
\midrule
\multirow{7}{*}{Human--human} & HIIv2~\citep{haroon2022hiiv2} & 2,529 & Two-person and group interactions \\
 & HII~\citep{tanisik2016hii} & 1,471 & Two-person interactions \\
 & PIC2.0~\citep{liu2021pic2} & 1,108 & Nonverbal interaction in crowds \\
 & AVA~\citep{gu2018ava} & 239 & Film frames, person-to-person actions \\
 & NVI~\citep{wei2024nvi} & 113 & Nonverbal interaction detection \\
 & Handover~\citep{carfi2019handover} & 66 & Object hand-over between people \\
 & UT-Interaction~\citep{ryoo2010utinteraction} & 45 & Surveillance-view interactions \\
\midrule
\multirow{6}{*}{Procedural} & NurViD~\citep{hu2023nurvid} & 401 & Nursing procedures \\
 & MVOR~\citep{srivastav2018mvor} & 311 & Multi-view operating room \\
 & ExpVid~\citep{xu2026expvid} & 161 & Laboratory experiments \\
 & DaRA~\citep{niemann2026dara} & 100 & Warehouse and logistics activities \\
 & MM-OR~\citep{ozsoy2025mmor} & 52 & Robotic surgery operating room \\
 & EgoExOR~\citep{ozsoy2026egoexor} & 38 & Exocentric operating-room views \\
\midrule
\multirow{9}{*}{Industrial and agricultural} & TeaWeed-Action~\citep{han2025teaweeding} & 590 & Field work with hand tools \\
 & HA-ViD~\citep{zheng2023havid} & 384 & Assembly with tools and parts \\
 & ENIGMA-360~\citep{ragusa2026enigma360} & 229 & Industrial procedures \\
 & Construction-CMA~\citep{yang2023cma} & 81 & Construction site actions \\
 & IMPACT~\citep{wen2026impact} & 57 & Industrial manual tasks \\
 & InHARD~\citep{dallel2020inhard} & 28 & Human--robot assembly \\
 & CarDA~\citep{papoutsakis2024carda} & 18 & Car-part assembly \\
 & OpenMarcie~\citep{bello2026openmarcie} & 10 & Industrial maintenance \\
 & IndEgo~\citep{chavan2026indego} & 9 & Industrial exocentric views \\
\bottomrule
\end{tabular}}
\end{table}

\noindent\textbf{Model assistance.}\label{app:heir-quality} Annotation proposals combine multimodal labeling, entity localization, and interaction detection. On sources without native interactions, GPT-5.4~\citep{openai2026gpt54} provides initial labels; RF-DETR, LocateAnything, LLMDet, WeDetect, and Rex-Omni provide entity candidates~\citep{robinson2026rfdetr,wang2026locateanything,fu2025llmdet,fu2026wedetect,jiang2026rexomni}. SAM~3~\citep{carion2026sam3} refines boxes, SL-HOI~\citep{sun2026slhoi} proposes interaction pairs, and Gemini~3.6 Flash~\citep{google2026gemini36flash} labels merged candidates. For native-interaction sources, Gemini~3.1 Flash-Lite~\citep{google2026gemini31flashlite} proposes roles while preserving source boxes, nouns, and actions before human correction. Geometric duplicates merge at IoU $\geq0.95$; conflicting nouns remain for human adjudication.

\noindent\textbf{Human correction.} Thirteen annotators correct entity identities, boxes, nouns, directed action links, and roles, and ten reviewers cross-check the annotations. All 18,730 released images have complete human annotations of interactions and receive a second review by someone other than the initial annotator. Structural validation checks vocabulary membership, person actors, one role per actor--action--participant triple, and duplicates.


\noindent\textbf{Role definitions.} \texttt{target}: the entity directly acted upon, attended to, indicated, displayed, read, moved, or changed in this action; a racket being held and a box being opened are both targets. \texttt{instrument}: the tool, equipment, controller, or external implement the person uses to perform the action. \texttt{support}: the entity bearing body or object weight or supporting posture, standing, or riding; distinct from the action \texttt{support} (assisting a person), whose assisted person is the target. \texttt{source}: the origin that a person, entity, or content leaves, including removal, extraction, pouring out, departure, and the vessel drunk, eaten, or scooped from. \texttt{destination}: the endpoint that a person, entity, or content reaches, including placement, insertion, pouring into, hand-over, and entry. A container can therefore be a source or destination, depending on the action. \texttt{constraint}: the entity limiting, obstructing, fixing, or restricting movement or state. A role describes the entity's function in this action, not its noun category. Fixed-role actions take one role by definition (\texttt{hold}, \texttt{grasp}, \texttt{touch}, \texttt{wear}: \texttt{target}; \texttt{sit}, \texttt{lie}, \texttt{stand}, \texttt{ride}: \texttt{support}); other actions distribute their participants over roles as the event requires.

\noindent\textbf{Participant identity.} People, objects, and body regions have image-local entity identities. A person participating in another actor's action retains the same identity across events. In the benchmark definition, an actor's affected body region is localized separately. Merely wearing or holding an object does not make the whole actor a self-directed participant.

\noindent\textbf{Class support.}\label{app:heir-additional-stats} Of 2,972 observed action--noun--role classes, 504 have at least ten training examples, 2,323 have one to nine, and 145 have no training examples. The corresponding test-supported counts are 424, 547, and 133, with 10,374, 1,201, and 208 test relations. The six roles have 51,142 target, 3,311 instrument, 12,036 support, 761 source, 1,081 destination, and 472 constraint relations across the release.

\section{CoRISP Parameterization and Training}
\label{app:rift-details}
\noindent\textbf{Visual inputs.} Frozen H-DETR~\citep{jia2023hdetr} with a Swin-L backbone~\citep{liu2021swin} supplies proposals. DINOv3 ViT-L/16 features pass through a pretrained DINO.txt vision head for text alignment~\citep{simeoni2025dinov3,jose2025dinotxt}; both components are frozen. A trainable projection maps person-proposal features into queries processed with image tokens by this vision head. Gradients pass through the head to the projection. Person, entity, union-region, geometry, and global features initialize 384-dimensional interaction states.

\noindent\textbf{Local distribution.} For a visible candidate $i=(h,j,v)$, semantic action logit $\zeta_i$, interaction residual $\eta_i$, role residual $d_{i,r}$, and learned scale $\lambda$, the logits are
\begin{equation}
 z_{i,0}=0,\qquad z_{i,r}=\zeta_i-\log|\mathcal R_v|
 +\lambda(\eta_i+d_{i,r}-\bar d_i),\quad
 \bar d_i=|\mathcal R_v|^{-1}\sum_{r\in\mathcal R_v}d_{i,r}.
\end{equation}
The categorical distribution is $p_i=\operatorname{softmax}(z_i)$. At zero residual, $q_i=\operatorname{sigmoid}(\zeta_i)$ independently of role count. The conditional role mean uses $\epsilon=10^{-30}$. Event messages are averaged with $q_i$; entity messages first average over actions and then across incoming pairs with weights $1-\prod_v(1-q_{hj,v})$. Entity states with zero incoming mass are retained. Positive-role probabilities are multiplied by detector-confidence and native visibility factors, yielding $\widetilde p_i(r)$ and $\widetilde p_i(0)=1-\sum_r\widetilde p_i(r)$.

\noindent\textbf{Role pooling.} The neighborhoods are $\mathcal N_{\rm event}(i)=\{(h,j',v)\}$, $\mathcal N_{\rm pair}(i)=\{(h,j,v')\}$, and $\mathcal N_{\rm entity}(i)=\{(h',j,v')\}$ over valid indices. Let $w_{i,r}=\operatorname{stopgrad}(\widetilde p_i(r))$. For query role $r$ and source role $r'$, the summary is
\begin{equation}
 \boldsymbol\mu^g_{i,r'\mid r}=
 \frac{\sum_{i'\in\mathcal N_g(i)}w_{i',r'}\mathbf v_{i',r'}
 -\mathbf1\{r'=r\}w_{i,r}\mathbf v_{i,r}}
 {\sum_{i'\in\mathcal N_g(i)}w_{i',r'}
 -\mathbf1\{r'=r\}w_{i,r}}.
 \label{eq:rift-context}
\end{equation}
The neighborhood includes $i$ before subtraction. Zero-mass summaries are masked; other denominators are floored numerically. Scaled dot-product attention~\citep{vaswani2017attention} uses local features to query summaries augmented with role and relation embeddings. A second attention fuses neighborhood messages. Refinement uses $\delta_{i,r}=4\tanh\beta\,\tanh\Delta_{i,r}$, where $\Delta_{i,r}$ is predicted from the bound state and $\beta$ is initialized to zero. The same detached weights pool bound states within each event and role; adding event and action--role features followed by layer normalization yields $\mathbf U^{\rm role}_{h,v}$.

\noindent\textbf{Count potentials.} A linear head on the normalized event state predicts $B(k)=f_B(\mathbf e_{h,v})_k-f_B(\mathbf e_{h,v})_0$. For role count embeddings $\boldsymbol\kappa_t$, $t\in\{0,1,2\}$, and learned projections of width $d_G$, let $R=|\mathcal R_v|$ and $\mathbf U_r$ denote one role-context row:
\begin{align}
 \mathbf f_r(c_r)&=\operatorname{LN}\bigl(\phi_U(\mathbf U_r)+\boldsymbol\kappa_{c_r}
 +\phi_U(\mathbf U_r)\odot\boldsymbol\kappa_{c_r}\bigr),\\
 g(\mathbf c)&=d_G^{-1/2}\phi_E(\mathbf e_{h,v})^\top
 \phi_S\!\left(R^{-1/2}\sum_r\mathbf f_r(c_r)\right),\\
 G(\mathbf c)&=4\tanh\beta_G\,\tanh\bigl(g(\mathbf c)-g(\mathbf0)\bigr).
\end{align}
The scale $\beta_G$ is initialized to zero. Assignment-dependent inputs to $B$ and $G$ are counts; candidate-specific compatibility enters through the contextual unaries.

\noindent\textbf{Dynamic program.}\label{app:rift-dp} Set disallowed positive unary weights to $-\infty$ without changing $\ell_m(0)$. Initialize $D_0(0,\mathbf0)=0$ and all other states to $-\infty$. Let $T_r$ increment the role-$r$ count, saturating at two. Then
\begin{align}
 A_{m+1}(k,\mathbf c)&=\operatorname{LSE}_{\substack{r\in\mathcal R_v,\ \mathbf c':\\T_r(\mathbf c')=\mathbf c}}
 [D_m(k-1,\mathbf c')+\ell_{m+1}(r)],\\
 D_{m+1}(k,\mathbf c)&=\operatorname{LSE}\bigl(D_m(k,\mathbf c)+\ell_{m+1}(0),\,A_{m+1}(k,\mathbf c)\bigr).
\end{align}
The predecessor sum includes only valid states; both counts $1$ and $2+$ can precede a saturated count. Applying the terminal potentials gives Equation~\eqref{eq:rift-partition}. The dense recurrence takes $O(M^2R3^R)$ time and $O(M3^R)$ rolling-table storage, excluding backward intermediates. Removing unreachable states is exact under the same support. For V-COCO, role-specific null candidates represent missing fillers and enter the same event-level inference.

\noindent\textbf{Matching and loss.}\label{app:rift-training} HEIR matches people one-to-one at IoU $\geq0.5$; participant proposals require the annotated noun and IoU $\geq0.5$. Compatible complete assignments use each proposal at most once and deduplicate matching paths. Ambiguous person targets and positives without a compatible proposal assignment are ignored during training, not removed from evaluation. Unobserved actions are treated as negatives only where the source annotation defines the corresponding supervision as complete. Cropping away a participant makes the positive event unknown rather than a smaller complete event.

We apply focal modulation~\citep{lin2017focal} to the set likelihood. For $L_y=\mathcal L_{\rm set}(y)$, supervised events $\mathcal O$ in a local microbatch, $W$ distributed ranks, and positive participant-group counts $N_w^+$, the loss is
\begin{equation}
 \mathcal L=
 \frac{\sum_{y\in\mathcal O}0.5(1-e^{-L_y})^{0.1}L_y}
 {\max(1,W^{-1}\sum_{w=1}^{W}N_w^+)}.
\end{equation}
Gradients are averaged across ranks; normalization is per microbatch. Visual computations use BF16 autocast, and log-space DP uses FP32. HEIR also evaluates structured-potential projections in FP32; V-COCO retains autocast for those projections. Frozen modules are excluded from trainable-parameter counts but remain part of inference.

\section{Evaluation and Supporting Results}
\label{app:experimental-details}\label{app:heir-metrics}
\noindent\textbf{HEIR relations.} Role mAP averages AP over 1,104 test-supported action--noun--role classes. HOI mAP instead uses 1,057 action--noun classes, retaining the maximum role score for each person--participant--action tuple. This common projection is applied to every model. For CoRISP, the benchmark score is $\max_r\pi_m(r)$, while the model's interaction probability is $q_m^{\rm set}=\sum_r\pi_m(r)$. Rare, non-rare, and unseen classes have 1--9, at least 10, and zero training instances, respectively.

\noindent\textbf{V-COCO.} V-COCO~\citep{gupta2015vcoco} annotates COCO images~\citep{lin2014coco}, with 2,533 training, 2,867 validation, and 4,946 test images. We train on trainval and evaluate all test images. Role AP under Scenarios~1 and~2 is denoted by $\mathrm{AP}_{\mathrm{role}}^{\mathrm{S1}}$ and $\mathrm{AP}_{\mathrm{role}}^{\mathrm{S2}}$, respectively. Our re-evaluations average the 24 action--role classes remaining after excluding \texttt{point}. Both scenarios use the authors' matching rules\footnote{\url{https://github.com/s-gupta/v-coco}}: a correct detection requires the action and role labels and IoU $\geq0.5$ for the person and each evaluated visible filler. For an absent ground-truth filler, S1 requires a null prediction, whereas S2 ignores its object localization.

\noindent\textbf{HEIR baseline sets.} Same-noun prediction boxes are clustered at IoU $\geq 0.7$ against cluster representatives, and duplicate edges retain their maximum score. For each predicted person identity and action, edges are sorted by confidence. Every nonempty prefix forms a set hypothesis, scored by its minimum member confidence. Baselines use this decoder, retaining at most 100 sets per image without an additional relation cap or a member-score threshold. Both baseline and CoRISP sets are evaluated with one shared, noun-compatible, one-to-one entity correspondence per image at IoU $\geq 0.5$. The test set contains 9,955 annotated sets across 101 actions; 85.5\% have one member.

\noindent\textbf{Shared entity correspondence.} Each predicted entity appearing in a submitted set receives the maximum score of any set containing it as actor or participant. Entities are processed in descending order of this priority, then descending entity confidence, then ascending entity ID. Each is assigned to the unused ground-truth entity with the same noun and greatest IoU, provided IoU $\geq0.5$; equal IoUs are resolved by ascending ground-truth ID. The resulting correspondence is reused across all events in the image. Action and role agreement are checked when matching complete sets.

\noindent\textbf{Set AP.} Within each action, hypotheses are ordered by decreasing set score, with image and hypothesis IDs resolving ties for matching. A true positive must have every endpoint matched and exactly equal an unused annotated event's entity--role set. Each annotated event can be used once; unmatched hypotheses and subsequent duplicates count as false positives. Equal-score outcomes are grouped before computing precision and recall. AP integrates the all-point interpolated precision envelope, and Set mAP averages AP over the 101 test-supported actions.

\noindent\textbf{Structure-specific Set AP.} Multi restricts both ground-truth sets and submitted hypotheses to at least two entity--role members; Repeat requires at least two distinct entities with the same role. These strata contain 1,441 and 874 ground-truth sets over 62 and 34 actions, respectively. The 100-set budget is applied before stratum filtering, and correspondence is recomputed within each stratum. These scores measure exact recovery conditional on the specified member structure. Shared evaluates all 3,217 person--action sets over 64 actions in the 534 images containing cross-actor participant sharing. Full-test Set mAP and these structural breakdowns together describe overall and composition-specific performance.

\noindent\textbf{CoRISP set prediction.} CoRISP submits sets directly from its event potentials. A max-plus dynamic program finds the highest-scoring assignment for each reachable total-count and saturated-role-count state. Each state contributes one winner. We rank these winners by their exact probability under the full distribution, retain at most $K=8$ per person--action event, and apply the same 100-set image budget. The resulting alternatives represent different count states. This readout is fixed across CoRISP evaluations and component ablations.

\noindent\textbf{V-COCO sets.} Our complete-set metric extends V-COCO evaluation to joint slot recovery over 21 role-bearing actions, excluding \texttt{point}. Predicted person boxes are clustered at IoU $\geq 0.7$ against fixed representatives, in descending order of their maximum role score. Each cluster--action pair forms one hypothesis using the highest-scoring prediction for each native role slot, scored by the minimum slot confidence. Every slot requires an explicit prediction, including an explicit null for a missing filler. No confidence threshold or additional output cap is applied. Matching follows the native person-matching convention and S1/S2 missing-filler rules, requiring all role slots to be correct. Set mAP averages interpolated AP over actions, grouping tied scores. Dual averages the three two-slot actions (\texttt{hit}, \texttt{eat}, and \texttt{cut}), including their absent-filler cases under S1/S2.

\noindent\textbf{Training.} CoRISP keeps the proposal detector and visual--semantic encoders frozen while training its feature projections, recurrent interaction field, context aggregation, and set potentials. The complete model has 9.92M trainable parameters on HEIR and 9.91M on V-COCO, excluding frozen components. The same architecture uses six functional roles on HEIR and native \texttt{obj}/\texttt{instr} slots on V-COCO. Both datasets use a 30-epoch training schedule. HEIR checkpoints are selected by validation Role mAP. V-COCO uses 5,267 eligible trainval images with visible-participant or missing-filler supervision and evaluates all 4,946 test images.

\noindent\textbf{Optimization.} The V-COCO model and HEIR component runs use AdamW with learning rate $10^{-4}$, weight decay $10^{-4}$, and gradient-norm clipping at 0.1. The learning rate is multiplied by 0.2 after epoch 20. Two accumulated microbatches give an effective batch size of 16; the random seed is 42. V-COCO uses the final epoch, and the HEIR component runs use validation Role mAP for checkpoint selection.

\noindent\textbf{Component protocol.}\label{app:component-protocol}
Each ablation is trained independently with the same optimization schedule, checkpoint criterion, set prediction rule, and output budget as the full model.

\noindent\textbf{Model configurations.} SOV-STG Swin-L uses resolution 384 on HEIR. Its V-COCO model and SOV-STG R101 on HEIR are trained with official code (T); SOV-STG R101 on V-COCO uses the released SOV-STG-L weights (W). MUREN and SOV-STG-S use ResNet-50~\citep{he2016resnet}, while GEN-VLKT-S/L use ResNet-50/101. GroupHOI-S uses ResNet-50 with CLIP-B/16~\citep{radford2021clip}, InCoM-Net uses ResNet-50 with CLIP-L/14, and SL-HOI uses DINOv3-L/16. SOV-STG-VLA-S uses ResNet-50 with BLIP-2~\citep{li2023blip2} on HEIR and CLIP-B/32 on V-COCO; its HEIR parameter count does not describe the V-COCO configuration. The published HOI-IDiff result lists Deformable DETR~\citep{zhu2021deformable}; UniHOI uses VQGAN~\citep{esser2021taming} and Llama3-8B~\citep{grattafiori2024llama3} with 550K additional image--text pairs. Score superscripts W/T/A/R indicate released weights, our training with official code, adapted official code, and our reimplementation, respectively.

UniHOI's released materials do not provide its task checkpoint and complete HOI inference pipeline; HOI-IDiff does not specify the exact task detector checkpoint and the full diffusion-training and role-output configuration. We retain their published V-COCO results under the cited protocols.

\noindent\textbf{Comparison scope.} Baselines retain the encoders and pretraining indicated in the main tables. Dagger-marked HEIR baselines use the shared inventory within their standard inference pipelines; other baselines retain their native support. All set outputs share the same 100-set budget. Their set outputs share the 100-set budget. Frozen upstream components are included in the system, but not in trainable-parameter counts. Native V-COCO role averaging and complete-set evaluation remain distinct from HEIR's noun-qualified, image-level identity criterion.

\noindent\textbf{Number of set hypotheses.} Table~\ref{tab:rift-decoding} holds the primary HEIR checkpoint fixed. Both settings score complete assignments using the event potentials: one returns the best nonempty assignment, while the other retains up to eight winners from different count states.

\begin{table}[t]
\centering
\caption{\textbf{Effect of the set hypothesis budget} (Set mAP, \%). The CoRISP checkpoint is fixed. $K$ counts state winners; both settings retain at most 100 sets/image.}
\label{tab:rift-decoding}
\small
\setlength{\tabcolsep}{5pt}
\begin{tabular}{@{}lr@{}}
\toprule
\textbf{Readout} & \textbf{Full}$\uparrow$ \\
\midrule
Single best nonempty assignment & \underline{17.56} \\
State-wise MAP, $K=8$ & \textbf{18.04} \\
\bottomrule
\end{tabular}
\end{table}

\noindent\textbf{Action--cardinality oracle.} For each annotated person--action event $e$, we reveal its true action and member count $k_e^\star=|S_e^\star|$. Predicted boxes are first clustered with same-noun IoU $\geq0.7$. People are matched greedily by IoU, and each ground-truth event then receives a noun-compatible, one-use participant matching at IoU $\geq0.5$. Let $\sigma_e$ rank that event's predicted entity--role edges by confidence. The oracle recovery rate is
\begin{equation}
R_{k^\star}=\frac{1}{|\mathcal D|}\sum_{e\in\mathcal D}\mathbf{1}\!\left[\operatorname{Top}_{k_e^\star}(\sigma_e)=S_e^\star\right].
\label{eq:cardinality-oracle}
\end{equation}
Unmatched events count as failures. Coverage $C$ is the fraction whose subject and every participant admit the required box-and-noun matching. Both $C$ and $R_{k^\star}$ are event-averaged recalls with the true action and set size supplied. Matching is performed within each event for this diagnostic, following the procedure above.

\begin{table}[!tp]
\centering
\caption{\textbf{Action--cardinality oracle at 100 relations per image} (\%). $C$: box-and-noun coverage; $R$: exact top-$k^\star$ recovery with the true action and count supplied. Denominators include all 1,441 Multi and 874 Repeat events. Denominators include all 1,441 Multi and 874 Repeat events.}
\label{tab:cardinality-oracle}
\begingroup
\small
\setlength{\tabcolsep}{3pt}
\begin{tabularx}{\linewidth}{@{}>{\raggedright\arraybackslash}X *{4}{>{\centering\arraybackslash}p{0.12\linewidth}}@{}}
\toprule
\textbf{Method} & \multicolumn{2}{c}{\textbf{Multi}} & \multicolumn{2}{c}{\textbf{Repeat}} \\
\cmidrule(lr){2-3}\cmidrule(l){4-5}
 & $C\uparrow$ & $R\uparrow$ & $C\uparrow$ & $R\uparrow$ \\
\midrule
\multicolumn{5}{@{}l}{\emph{Visual models}} \\
QPIC R50 \citep{tamura2021qpic} & 28.94 & 14.99 & 31.69 & 16.25 \\
QPIC R101 \citep{tamura2021qpic} & 31.58 & 15.75 & 33.98 & 16.48 \\
MUREN \citep{kim2023muren} & 34.91 & 16.79 & 40.16 & 17.85 \\
SOV-STG-S \citep{chen2025sovstg} & 42.19 & 17.49 & 49.08 & 18.88 \\
SOV-STG Swin-L \citep{chen2025sovstg} & \textbf{51.21} & \underline{24.22} & 56.06 & 24.94 \\
SOV-STG R101 \citep{chen2025sovstg} & 43.44 & 18.46 & 49.31 & 18.76 \\
PViC R50 \citep{zhang2023pvic} & 37.82 & 17.77 & 46.11 & 21.85 \\
PViC Swin-L \citep{zhang2023pvic} & 47.47 & 22.55 & \underline{60.41} & \underline{26.89} \\
\midrule
\multicolumn{5}{@{}l}{\emph{Vision--language models}} \\
GEN-VLKT-S \citep{liao2022genvlkt} & 44.97 & 16.10 & 49.77 & 17.96 \\
GEN-VLKT-L \citep{liao2022genvlkt} & 48.09 & 19.08 & 52.63 & 20.14 \\
SOV-STG-VLA-S \citep{chen2025sovstg} & \underline{49.13} & 15.61 & 53.32 & 16.93 \\
RLIPv2 Swin-T \citep{yuan2023rlipv2} & 41.71 & 21.58 & 49.31 & 25.06 \\
RLIPv2 Swin-L \citep{yuan2023rlipv2} & 48.30 & 23.94 & 54.58 & 25.51 \\
GroupHOI-S \citep{hong2025grouphoi} & 40.60 & 12.28 & 47.94 & 13.39 \\
InCoM-Net \citep{seo2026incom} & 31.51 & 17.14 & 41.42 & 21.62 \\
SL-HOI \citep{sun2026slhoi} & 25.75 & 6.87 & 30.09 & 6.29 \\
\midrule
CoRISP & 48.92 & \textbf{27.48} & \textbf{62.01} & \textbf{33.41} \\
\bottomrule
\end{tabularx}
\endgroup
\end{table}

\noindent\textbf{Benchmark diagnostics.}\label{app:benchmark-diagnostics} Figure~\ref{fig:heir-diagnostics} compares all 16 HEIR baselines in Table~\ref{tab:mainresults}. Action-balanced Role AP first averages supported noun--role classes within each action, then averages over the same 101 actions as Set mAP. Under equal action weights, 41 of 120 model pairs reverse order; ten of 55 reverse among shared-inventory baselines. Multi-role-pair Role AP weights each of the 47 test-supported action--noun pairs with multiple roles equally, averaging its role AP values first. Both analyses use the full test predictions.

For the pair-balanced comparison in Figure~\ref{fig:heir-diagnostics}(a), HOI AP averages one AP per action--noun pair, while Role AP first averages that pair's supported role classes and then averages pairs. The multi-role group contains 47 pairs, 94 role classes, and 630 relations; the single-role group contains 1,010 pairs and 11,153 relations. Their AP gap measures sensitivity to role labels under equal pair weights.

Edge coverage requires every ground-truth member to occur in the union of retained hypotheses; exact recovery requires one hypothesis to equal the entire set. Under the same 100-set budget, the latter recovers 54.86\%, 56.49\%, 53.85\%, 61.81\%, and 53.58\% of edge-covered multi-participant events for PViC R50, PViC Swin-L, GroupHOI-S, InCoM-Net, and SL-HOI. Every covered but unrecovered event has an extra member in each covering prefix. Each recall is conditioned on that model's own covered events.

\noindent\textbf{Classification with correct grounding.}\label{app:conditional-grounding} We compare CoRISP with SL-HOI, the strongest relation-level baseline under this common-grounding analysis, on a shared set of person--participant pairs. Each model's predicted pairs are matched to annotated pairs once, first maximizing the number with person and participant IoU $\geq0.5$ and the correct foreground noun, then maximizing the sum of their minimum endpoint IoUs. Matching uses geometry and nouns only. The intersection contains 10,383 pairs from 2,495 images, including 4,568 pairs without a positive annotated relation. Both models retain their predicted scores; actions and roles are predicted over identical pairs and supervision scopes. AP averages classes with at least one positive in this shared population. CoRISP and SL-HOI obtain 63.05/50.97 conditional HOI mAP over 562 action--noun classes and 61.50/50.26 conditional Role mAP over 584 action--noun--role classes. Among the 214 supported rare role classes, the scores are 58.68/47.28. These values measure classification on common correctly grounded candidates; the features and detection confidences remain those of each model.

\begin{table}[!tp]
\centering
\caption{\textbf{Published V-COCO role AP} (\%, S1/S2). Source averaging conventions differ; scores are therefore unranked. R/C/D: ResNet/CLIP/DINOv3; DDETR: Deformable DETR.}
\label{tab:vcoco-reported}
\begingroup
\small
\setlength{\tabcolsep}{4pt}
\begin{tabularx}{\linewidth}{@{}>{\raggedright\arraybackslash}X l r@{}}
\toprule
\textbf{Method} & \textbf{Encoder(s)} & \textbf{Reported role AP}$\uparrow$ \\
\midrule
\multicolumn{3}{@{}l}{\emph{Visual models}} \\
QPIC \citep{tamura2021qpic} & R50 & 58.8/61.0 \\
QPIC \citep{tamura2021qpic} & R101 & 58.3/60.7 \\
MUREN \citep{kim2023muren} & R50 & 68.8/71.0 \\
SOV-STG-S \citep{chen2025sovstg} & R50 & -- \\
SOV-STG R101 \citep{chen2025sovstg} & R101 & 63.9/65.4 \\
PViC \citep{zhang2023pvic} & R50 & -- \\
PViC \citep{zhang2023pvic} & Swin-L & 64.1/70.2 \\
\midrule
\multicolumn{3}{@{}l}{\emph{Vision--language models}} \\
GEN-VLKT-S \citep{liao2022genvlkt} & R50 & 62.41/64.46 \\
GEN-VLKT-L \citep{liao2022genvlkt} & R101 & 63.58/65.93 \\
SOV-STG-VLA-S \citep{chen2025sovstg} & R50 + C-B/32 & 63.8/65.7 \\
RLIPv2 \citep{yuan2023rlipv2} & Swin-T & 68.8/70.8 \\
RLIPv2 \citep{yuan2023rlipv2} & Swin-L & 72.1/74.1 \\
GroupHOI-S \citep{hong2025grouphoi} & R50 + C-B/16 & 65.0/66.0 \\
InCoM-Net \citep{seo2026incom} & R50 + C-L/14 & 73.6/75.4 \\
SL-HOI \citep{sun2026slhoi} & D-L/16 & -- \\
\bottomrule
\end{tabularx}
\endgroup
\end{table}

\noindent\textbf{Scoring convention.} Identical GEN-VLKT-S predictions score 62.48/64.53 over 25 action--role classes and 65.05/67.19 over the 24 classes excluding \texttt{point}. The change follows solely from the averaging convention. All re-evaluated $\mathrm{AP}_{\mathrm{role}}$ scores in Table~\ref{tab:mainresults} use the latter convention; literature-only rows retain source averaging conventions and are displayed separately.

\end{document}